\documentclass{svproc}

\usepackage{url}

\usepackage{booktabs} 

\usepackage{cite}
\usepackage{amsmath,amssymb,amsfonts}
\usepackage{graphicx, enumitem, booktabs}
\usepackage{multirow}
\usepackage{textcomp}
\usepackage{subfigure}
\usepackage[ruled,vlined,linesnumbered,longend]{algorithm2e}
\usepackage[table]{xcolor}
\usepackage{tikz}

\begin{document}
\mainmatter              
\title{Age-Friendly Route Planner: Calculating Comfortable Routes for Senior Citizens}
\titlerunning{Age-Friendly Route Planner}
\author{Andoni Aranguren\inst{1} \and Eneko Osaba\inst{1,2} \and \\ Silvia Urra-Uriarte\inst{1} \and Patricia Molina-Costa\inst{1}}
\authorrunning{Aranguren et al.} 
%
\tocauthor{Andoni Aranguren, Eneko Osaba, Silvia Urra-Uriarte, Patricia Molina-Costa}
\institute{TECNALIA, Basque Research and Technology Alliance (BRTA), Spain\\
\email{eneko.osaba@tecnalia.com},
\and
Corresponding Author}

\maketitle              

\begin{abstract}
The application of routing algorithms to real-world situations is a widely studied research topic. Despite this, routing algorithms and applications are usually developed for a general purpose, meaning that certain groups, such as ageing people, are often marginalized due to the broad approach of the designed algorithms. This situation may pose a problem in cities which are suffering a slow but progressive ageing of their populations. With this motivation in mind, this paper focuses on describing our implemented Age-Friendly Route Planner, whose goal is to improve the experience in the city for senior citizens. In order to measure the age-friendliness of a route, several variables have been deemed, such as the number of amenities along the route, the amount of \textit{comfortable} elements found, or the avoidance of sloppy sections. In this paper, we describe one of the main features of the Age-Friendly Route Planner: the \textit{preference-based routes}, and we also demonstrate how it can contribute to the creation of adapted friendly routes.
\keywords{Route Planning, Optimization, Smart Cities, Senior Citizens, Ageing People}
\end{abstract}
\section{Introduction}

Nowadays, route planning is a hot topic for both urban planners and the research community. The reason for this popularity can be broken down into two factors. On the one hand, due to their complexity, it is a tough challenge to solve this kind of problems. Hence, the inherent scientific appeal of these problems is irrevocable. On the other hand, the business benefits of efficient logistics and the social advantages that this would bring make addressing these problems of great interest for companies and civil servants. 

Evidence of this interest is the growing number of scientific publications that are added to the literature year after year \cite{precup2021nature,9781399,precup2020grey,osaba2021hybrid}. It is also interesting the growing number of open-source frameworks for route planning that can be found in the community, which can be used to solve routing problems of different kind. Examples of this kind of frameworks are Open Trip Planner (OTP, \cite{morgan2019opentripplanner}), Open-Source Routing Machine (OSRM, \cite{huber2016calculate}), or GraphHopper. 

Despite the large amount of research and developments made on the topic, routing algorithms and applications are usually developed for a general purpose, meaning that certain groups with mobility restrictions, such as ageing people, are often marginalised due to the broad approach of their designs. In most cases, routing algorithms aim to optimise efficiency factors, such as the traverse speed, distance, and public transport transfers. However, these factors can result in a route that is challenging for underrepresented groups with specific physical needs such as periodic resting, hydration to prevent heat strokes and incontinence. These groups, typically older people and people with with physical disabilities/ physical limitations, require route planners with a different approach that integrates accessibility factors.

In line with this, many European cities are experiencing a slow but progressive ageing of their populations, arising a considerable spectrum of new concerns that should be taken into account. For this reason, and as a result of these concerns, policy makers and urban planners are in a constant search of novel initiatives and interventions for enhancing the participation in the city life of senior citizens.

In this context, Artificial Intelligence (AI) has emerged as a promising field of knowledge area for dealing with ageing people related concerns. For this reason, a significant number of municipalities and cities have adopted AI solutions into their daily activity, implementing various systems for constructing innovative functionalities around, for example, mobility. However, the potential of AI for the development of innovative age-friendly functionalities remains almost unexplored, for example the development of age-friendly route planners. Few efforts have been made in this direction, such as the work recently published in \cite{abdulrazak2022toward} describing a preliminary prototype is described for planning public transportation trips for senior citizens.

With this motivation in mind, the main objective of this research is to present our developed Age-Friendly Route Planner, which is fully devoted to providing senior citizens with the friendliest routes. The goal of these routes is to improve the experience in the city for these ageing users. In order to measure this friendliness, several variables are considered, such as the number of amenities along the route, the number of elements that improve the comfortability of the user, or the consideration of flat streets instead of sloppy sections.

Specifically, in this paper we detail one of the main functionalities of our Age-Friendly Route Planner: the \textit{preference-based route planning}. Thanks to this functionality, adapted walking routes can be computed based on four weighted preferences inputted by the user and related to \textit{i)} the duration, \textit{ii)} the incline of the streets traversed, \textit{iii)} the amount of amenities found throughout the route, and \textit{iv)} the overall comfortability of the trip. The entire route planner has been implemented based on the well-known Open Trip Planner\footnote{ https://www.opentripplanner.org/}.

To properly develop this functionality, different real-world data have been used, and two ad-hoc data-processing engines have been implemented, namely, the Standardized Open Street Maps Enrichment Tool (SOET), and the Amenity Projection Tool (AOT). These tools, along with the preference-based route planning functionality, and the overall structure of the Age-Friendly Route Planner are described in detail along this paper. In addition, we show some solution examples in the city of Santander, Spain, to demonstrate the applicability of the planner we have developed.

It should be pointed out here that the \textit{preference-based route planning} functionality supposes a significant innovation for our Age-Friendly Route Planner regarding the vast majority of general-purpose route planners available in the literature, which do not compute this type of age-friendly routes. SOET and AOT also represent a remarkable contribution to this work, as they can be easily replicated in other route planners and Open Street Map (OSM, \cite{haklay2008openstreetmap}) based applications.

The structure of this paper is as follows. In the following Section \ref{sec:estr}, we detail the overall structure of the Age-Friendly Route Planner. In Section \ref{sec:data}, we describe the main data used by the planner to properly perform the functions that this paper focuses on. We also describe SOET and AOT in this section. In addition, in Section \ref{sec:routes}, we describe the preference-based route planning. In this section, we also introduce some examples of its applicability. Lastly, we finish this paper with conclusions and further work ( Section \ref{sec:conc}).

\section{The Age-Friendly Route Planner}
\label{sec:estr}

After analysing a significant amount of the most popular open source route planners (such as GraphHopper, OptaTrip, Traccar, OSRM or MapoTempo, among many others), we found that most of them are mainly designed for vehicle routing. This fact highlights the need for a solution that is primarily aimed at citizens. With this in mind, and as mentioned in the introduction, OTP has been selected as the framework for being used in this work. This fact does not imply that the advantages of other alternatives should be underestimated, but the main features, flexibility, and benefits that OTP offers to developers led us to choose it as an excellent platform for achieving the main objectives established. On closer examination, several important reasons led us to choose OTP as the base framework, which are the following:

\begin{itemize}
	\item It is fully open source, meaning that it can be fully customised to fulfill the research requirements.
	\item It works efficiently with widely known standards such as OSM or Geotiff (for defining city elevations).
	\item Being published in 2009, OTP is a platform with a long trajectory. Therefore it is very well documented and has a large and active community working on it. This facilitates the understanding of the framework.
	\item Both the API and the outcome JSON are fully customisable to the research requirements.
\end{itemize}

\begin{figure}[h!]
	\centering
	\begin{tabular}{cc}
		\includegraphics[width=0.95\columnwidth]{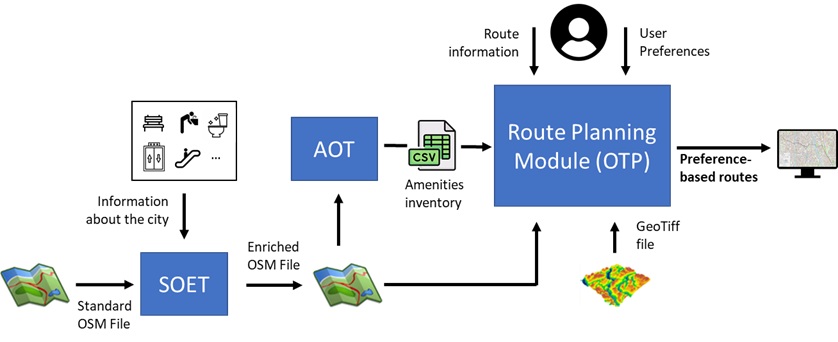}
	\end{tabular}
	\caption{Overall architecture of the Age-Friendly route planner.}
	\label{fig:OTP}
\end{figure}

As for the main structure of the Age-Friendly Route Planner, it has a central module coined as \textit{route planning module}, which is responsible for calculating routes using both the available data and the information entered by the user via API as input. In Figure \ref{fig:OTP} we represent the overall architecture of the Age-Friendly Route Planner, considering also the data needed for its correct use and the ad-hoc tools implemented for gathering the correct data.

Having said that, in order to properly contemplate all the requirements that the routing system should fulfill, the data sources described in the following section have been used. All these data sources are embedded in the OTP platform so that they can be taken into account in the correct planning of the routes.

\section{Data Sources and Data Processing Engines}
\label{sec:data}

In order to generate the required routes appropriately, the Age-Friendly Route Planner must build the corresponding street network. To do this, we need the corresponding OSM map file of the city in question; in the case of this study, Santander, Spain. The OSM format is fully compatible with OTP, which automatically consumes the files and builds the corresponding road network. 

In addition, this OSM file also takes into account important elements for the planner such as elevators, benches, fountains, toilets and automatic ramps. In line with this, it should be highlighted that the OSM files that can be openly obtained from open platforms are usually not as specific as we need them to be for our research. Open OSM files have proven to be very efficient for routing, but in terms of amenity related content, they are far from meeting the needs of this specific research. For this reason, we have developed an ad-hoc tool for enriching the Standardized OSM files. We have coined this tool as Standardized OSM Enrichment Tool, or SOET.
\subsection{Standardized OSM Enhanced Tool - SOET}

As explained earlier, for the Age-Friendly Route Planner, it is necessary to contemplate the amenities that are spread across the city, as this is a crucial factor for the success of the route planner. For this reason, the Santander's City Council, provided us with a series of files containing a list of amenities with the corresponding geospatial data. Among these amenities, we found benches, drinkable water sources, handrails and toilets. As the data was not provided in OSM format, it had to be pre-processed in order to be consumed. SOET was developed out of this motivation.

A data enrichment solution has been chosen over other options for this purpose, as data enrichment is in itself a crucial functionality for any project. Furthermore, as OSM is the basis for map creation in OTP, SOET provides a cascading effect for all applications based on the OSM file. Some examples are data visualisation in OTP and also the APT described later.

The required files for the Python-based SOET are the OSM file in which the data is stored and a \textit{.csv} file containing the amenities to be loaded. In the following Figure \ref{fig:SOETMap} we can see a clear example of an OSM file before enrichment (Figure \ref{fig:SOETMap}.a) and the result of applying SOET (Figure \ref{fig:SOETMap}.b). In this figure, we can see the newly added elements: Benches (pink dots), drinking fountains (blue dots) and garbage cans (red dots).

\begin{figure}[htbp]
	\centering
	\subfigure[]{\includegraphics[width=60mm]{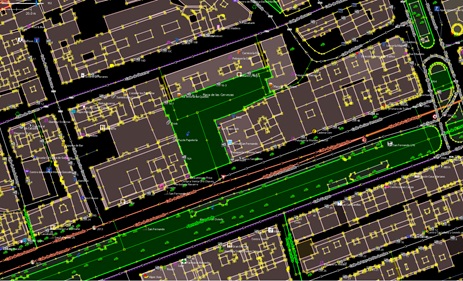}}
	\subfigure[]{\includegraphics[width=60mm]{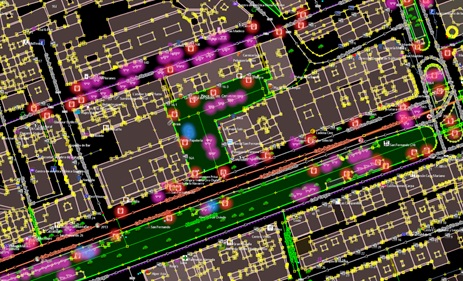}}
	\caption{A map excerpt of Santander before (a) and after (b) the SOET application.} 
	\label{fig:SOETMap}
\end{figure}

\subsection{Amenity Projection Tool - APT}
\label{sec:APT}

At this moment, four different types of amenities have been deemed in the route: public toilets, benches, handrails and drinking fountains. All these amenities have been extracted from the OSM maps. For this purpose, a new tool has been developed as part of this research. We have coined this tool as Amenity Projection Tool, or APT.

\begin{figure}[htbp]
	\centering
	\subfigure[APT criteria for considering amenities]{\includegraphics[width=40mm]{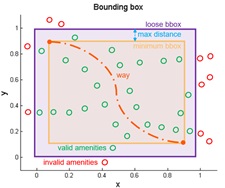}}
	\subfigure[How APT relates amenities and ways]{\includegraphics[width=40mm]{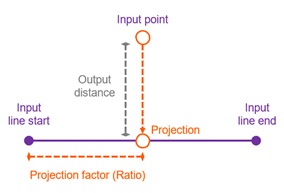}}
	\caption{Basic concepts of APT } 
	\label{fig:APTB}
\end{figure}

The APT was developed to correlate street segments (ways) and amenities (nodes). Thus, the AOT starts by reading the OSM file and then creates a bounding box that encloses each way. This bounding box is loose enough to enclose nearby nodes, and it is larger than the minimum bounding box by a user-defined maximum distance. So the bounding box is created to reduce the amount of correlation calculation which was $O(A_t*W_t)$ , where $A_t$ is the total amount of amenities and $W_t$ is the total number of ways. It reduces it to the bare minimum by adding a $O(W_t)$ pre-process resulting in a $O(A_p*W_p+A_t)$ = $O(A_p*W_p)$, where $A_p$ and $W_p$ are a subset of amenities and ways which are also less than or equal to $A_t$ and $W_t$. This approach has been chosen to ensure that the algorithm is efficient. In Figure \ref{fig:APTB}.a we represent this situation graphically, considering that the green amenities are close enough to the street, in order to correlate these amenities with the road.
After this first step, each amenity node (point) within the way bounding box is projected with an orthographic projection onto each path segment (lines). Several measures can be extracted from this projection, but only one is currently considered: the distance between the point and its projection. This distance indicates how far the amenity is from the segment. If it is equal to or less than the specified maximum distance parameter, then the amenity is added to its amenity type count. Figure \ref{fig:APTB}.b represents this situation.

After the execution of these two phases, all correlations among ways and amenities are stored in a \textit{.csv} which contains the identifications of all the ways and the number of amenities of each type that are within the parameterised maximum distance from that way.

\subsection{Elevation Data}
\label{sec:elevation}

In order to calculate friendly routes for senior citizens, the Age-Friendly route planner also considers the incline of the streets. This is a crucial aspect deeming that steep streets are usually preferred to be avoided and sometimes could create unwalkable routes for the older people. Consequently, a file containing the elevation of the city is compulsory. Luckily, OTP already allows the consumption of this information using the widely known GeoTIFF metadata standard. GeoTIFF permits the georeferenced of different information embedded into a \textit{.tif} file. With such a file, OTP can assign to a certain elevation to its corresponding street. For this purpose, the \textit{.tif} file has been obtained from the \textit{SRTM 90m Digital Elevation Database} open platform\footnote{https://cgiarcsi.community/data/srtm-90m-digital-elevation-database-v4-1/}.

\section{Building Preference-based routes in the Age-Friendly Route Planner}
\label{sec:routes}

In order to calculate routes based on user preferences, a functionality coined as \textit{Square Optimization} has been implemented in the Age-Friendly Route Planner. This kind of optimization allows the user to define four different preferences for the calculation of walking routes (whose sum must equal 100\%).

\begin{itemize}
	\item \textit{Slope}: this factor regards the incline of the route. The higher this factor, the flatter the routes calculated by the planner. The streets incline is calculated using the elevation data described in Section \ref{sec:elevation}.
	\item \textit{Duration}: this factor regards the duration of the route. The higher this factor, the shorter the routes calculated in terms of time.
	\item \textit{Amenities found along the route}: this factor considers the amenities described in the previous Section \ref{sec:APT}. In this case: benches, toilets and drinking water fountains. The higher this factor, the more amenities will be found along the route. In other words, a high value of this factor implies that the route planner will prioritize going through streets that contain these amenities.
	\item \textit{Comfortability factor}: the comfortability factor considers those elements that make the route more comfortable for the user. At the time of writing this paper, and because of a lack of additional data, only handrails have been included in this comfortability factor. In future stages of the Age-Friendly route planner, additional aspects such as shadows will be contemplated for this factor. Just like the amenities factor, the higher the comfortably factor, the more comfortable the routes will be.
\end{itemize}

\begin{figure}[htbp]
	\centering
	\subfigure{\includegraphics[width=50mm]{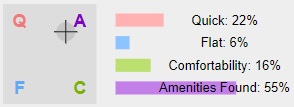}}
	\subfigure{\includegraphics[width=50mm]{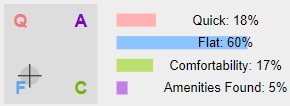}}
	\caption{Visual examples of the Square Optimization in the Age-Friendly route planner for walking routes.}
	\label{fig:widget}
\end{figure}

In order to demonstrate the applicability of this kind of routes, a testing purpose webpage has been deployed based on OTP. This page is fully accessible to any interested reader\footnote{https://afrp.santander.urbanage.digital.tecnalia.dev/}. In here the user is able to introduce their preferences using the interactive interface. We show in Figure \ref{fig:widget} two examples for this preference settings. Also, in this webpage, the user is able to choose routing options such as the origin and destination of the path.

\begin{figure}[htbp]
	\centering
	\subfigure[A preference-based route prioritizing quickness.]{\includegraphics[width=100mm]{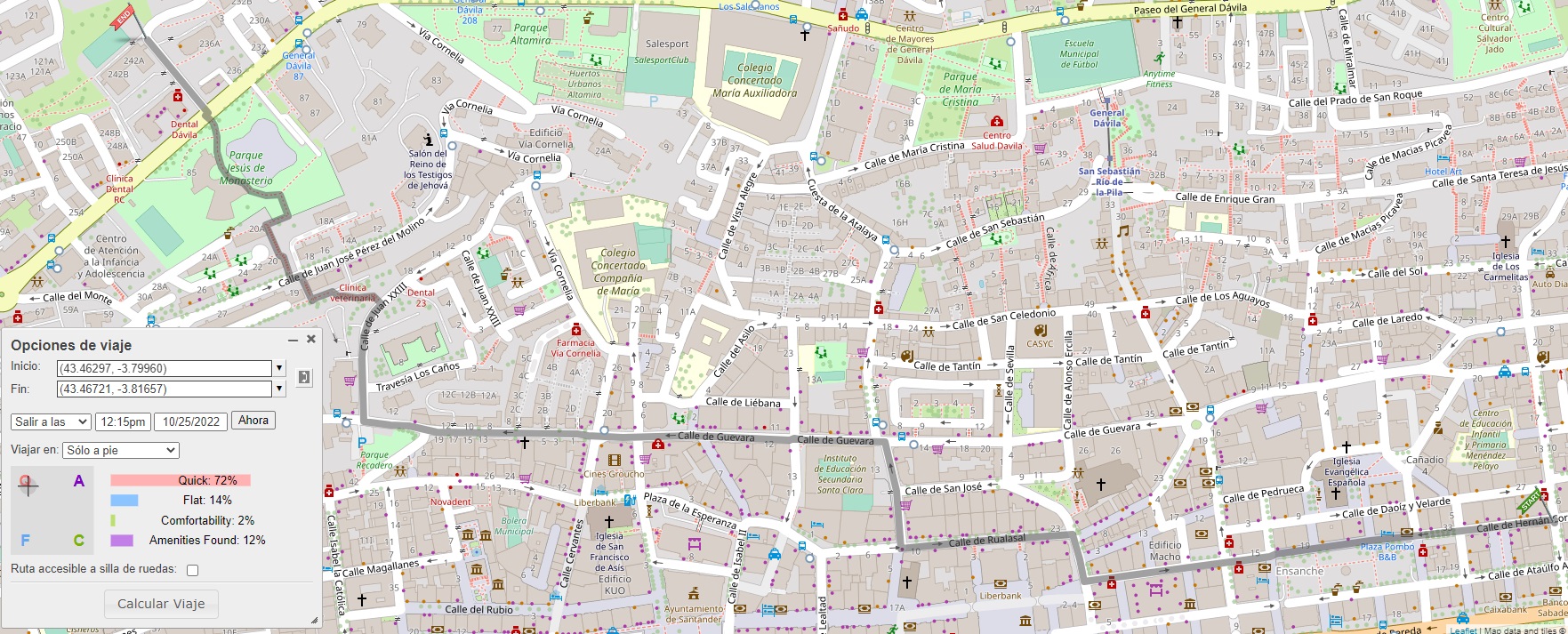}}
	\subfigure[A preference-based route prioritizing low-slope streets.]{\includegraphics[width=100mm]{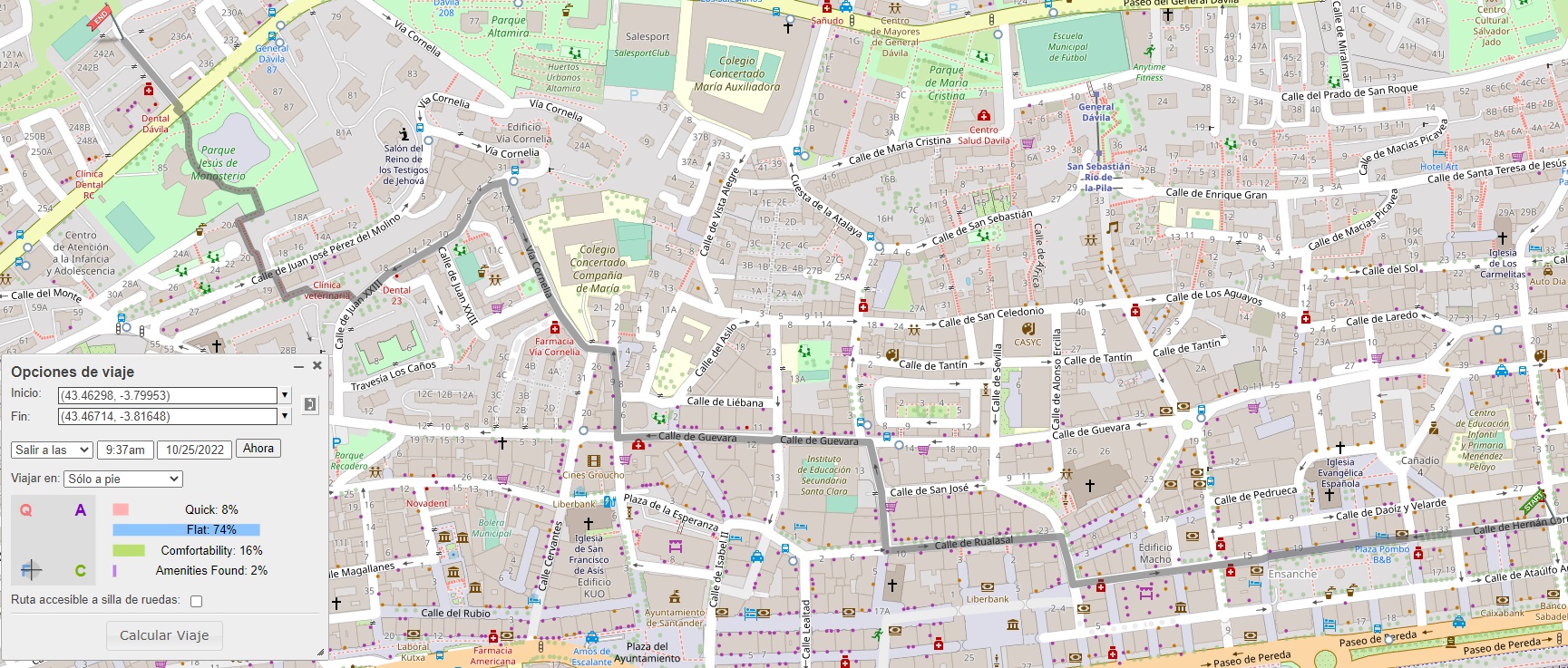}}
	\subfigure[A preference-based route prioritizing the finding of amenities.]{\includegraphics[width=100mm]{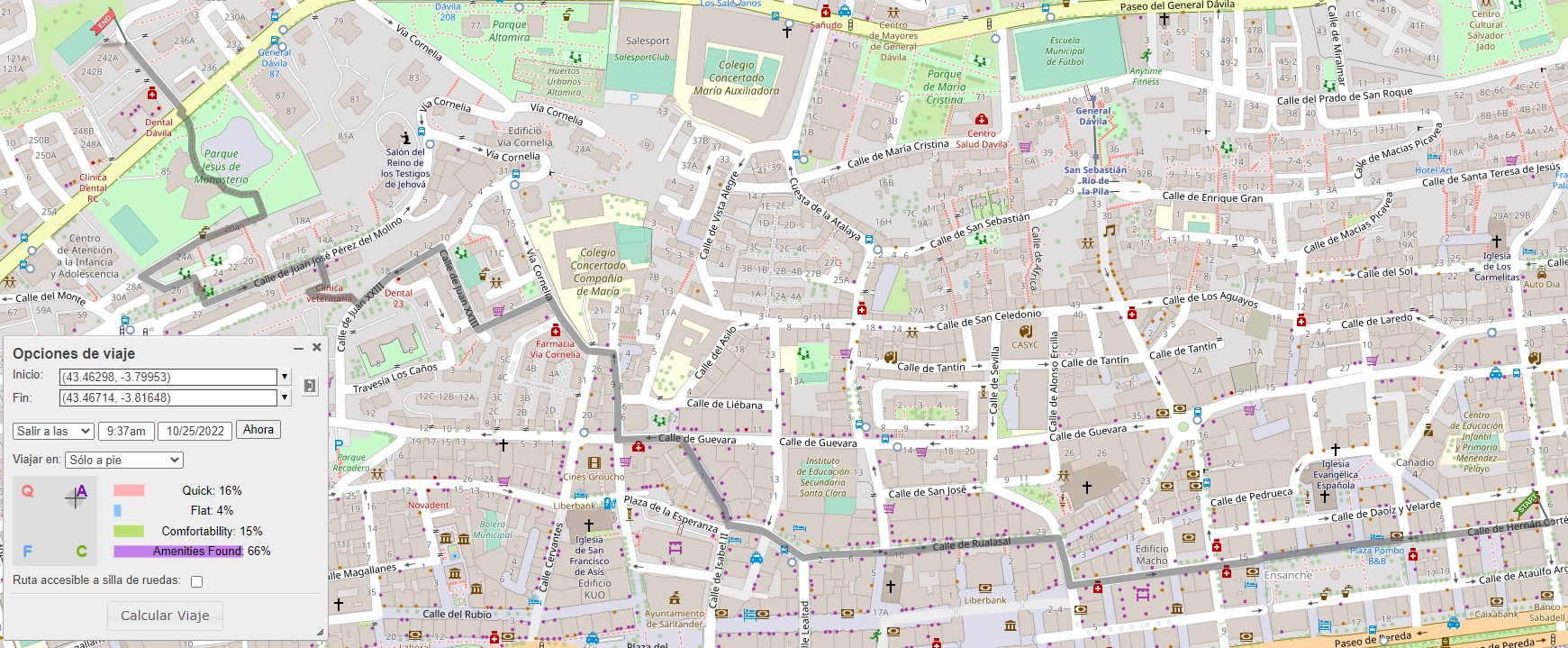}}
	\subfigure[A preference-based route prioritizing the comfortability of the route]{\includegraphics[width=100mm]{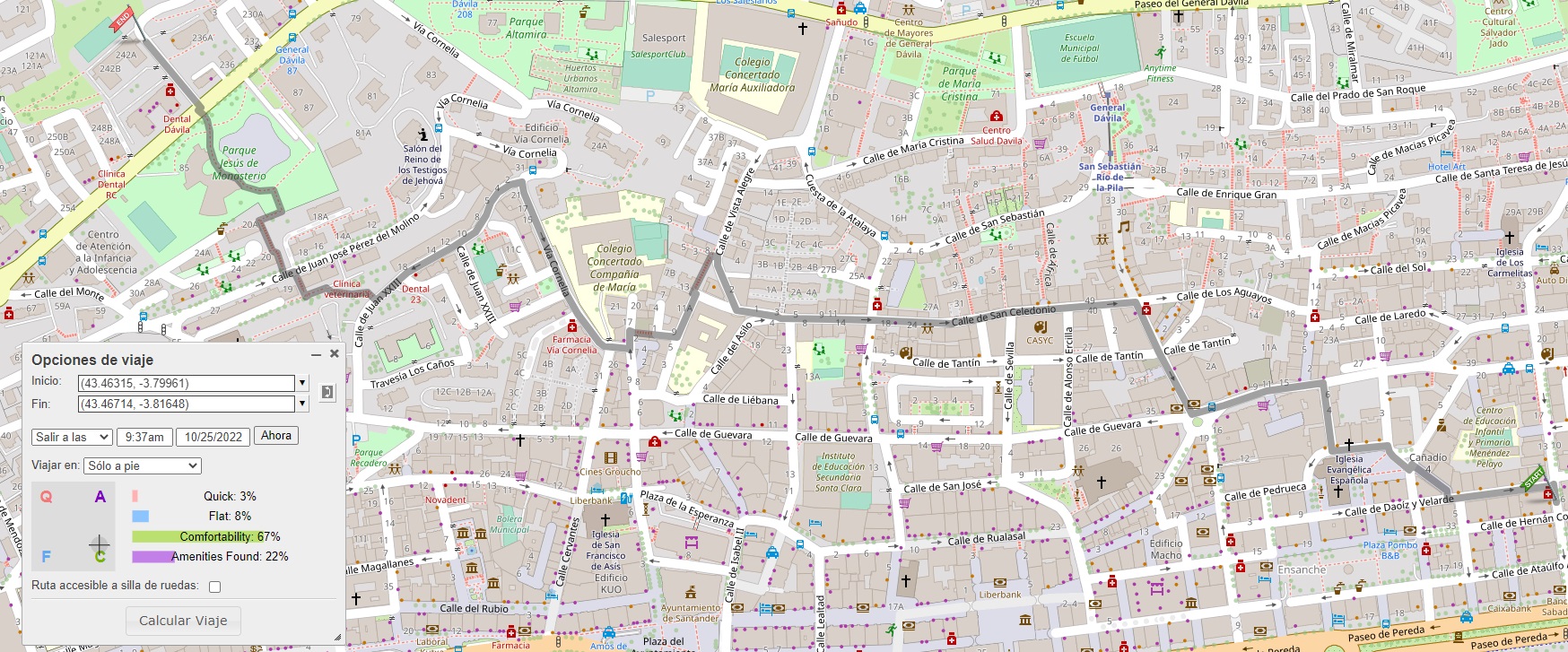}}
	\caption{Different examples demonstration the application of the preference-based walking routes functionality of the Age-Friendly Route Planner.}
	\label{fig:demo}
\end{figure}

Now we represent four different examples of walking routes, each one clearly prioritizing one of the deemed factors. As explained before, this demonstration is placed in the city of Santander, Spain; and in order to properly show the impact of the four factors, same origins and destinations are considered for each route. Thus, Figure \ref{fig:demo}.a represents a route which prioritizes quickness. Secondly, Figure \ref{fig:demo}.b depicts a path devoted to use low-incline routes. Meanwhile, Figure \ref{fig:demo}.c shows a route prioritizing the appearance of amenities along the route. Lastly, Figure \ref{fig:demo}.d represents a path which prioritizes comfortability, which means that it has handrails. Moreover, we depict in Table \ref{tab:summary} a summary of the features of each route, in order to clearly visualize the main characteristics of each path, and how the choosing of different preferences is crucial for building adapted routes.

\begin{table}[h!]
	\centering
	\resizebox{0.9\columnwidth}{!}{
		\begin{tabular}{l|c|c|c|c}
			\toprule[1.0pt]
			 & Route I-Fig \ref{fig:demo}.a &  Route II-Fig \ref{fig:demo}.b & Route III-Fig \ref{fig:demo}.c & Route IV-Fig \ref{fig:demo}.d\\
			\midrule[1.0pt]
			& \multicolumn{4}{c}{Preferences of the route} \\
			\midrule[1.0pt]
			Slope Factor & 14\% & \textbf{74\%} & 4\% & 8\% \\
			Duration Factor & \textbf{72\%} & 8\% & 15\% & 3\% \\
			Amenity Factor & 12\% & 2\% & \textbf{66\%} & 22\% \\
			Comfortability Factor & 2\% & 16\% & 15\% & \textbf{67\%} \\
			\midrule[1.0pt]
			& \multicolumn{4}{c}{Information of the route} \\
			\midrule[1.0pt]
			Incline & 487,3 & \textbf{447,3} & 504,5 & 514,5 \\
			Duration & \textbf{34min} & 38min & 37min & 38min\\
			Amenities & 101 & 88 & \textbf{161} & 120 \\
			Comfortable Elements & 40 & 46 & 60 & \textbf{65} \\
			\bottomrule[1.0pt]
			
		\end{tabular}
	}
	\caption{Parameters and information about the routes calculated. \textit{Incline}: sums the overall elevation-comfortability of the route (the less the better). \textit{Duration}: duration of the route. \textit{Amenities}: number of amenities found. \textit{Comfortable}: amount of comfortable elements.}
	\label{tab:summary}
\end{table}
\vspace{-1cm}

\section{Conclusions}
\label{sec:conc}

The application of routing algorithm to real-world situations has been a hot research topic in the last decades. As a result of this interest, the research carried out in this field is abundant. Despite this, routing algorithms and applications are usually developed for a general purpose, meaning that certain groups, such as ageing people, are often marginalized because of the broad approach of the designed algorithms. This situation may pose a problem in different parts of the world, such as Europe, in which many are experiencing a slow but progressive ageing on their populations, arising a considerable spectrum of new challenges and concerns that should be approached.

With this motivation in mind, this paper is focused in describing our own routing solution called Age-Friendly Route Planner. This planner is fully devoted to providing ageing citizens with the friendliest routes. The main objective of this route planner is to improve the experience in the city for senior people. To measure this friendliness, several variables have been taken into account, such as the number of amenities found along the route, the number of elements that improve the comfortability of the user along the path, the usage of urban infrastructures or avoiding sloppy sections.

Having shown and demonstrated one of the main functionalities of the Age-Friendly Route Planner, which is the \textit{preference-based route planning}, several research lines have been planned as future work. As a short term, we will implement further features on our route planner, such as in public transportation routes. or the consideration of people using wheelchair. As a long-term activity, we have planned to extend our Age-Friendly Route Planner to other European cities which might arise unique challenges.

\section*{Acknowledgments}

This work has received funding from the European Union’s H-2020 research and innovation programme under grant agreement No 101004590 (URBANAGE).

\bibliographystyle{IEEEtran}

\begin{thebibliography}{1}
	\providecommand{\url}[1]{#1}
	\csname url@samestyle\endcsname
	\providecommand{\newblock}{\relax}
	\providecommand{\bibinfo}[2]{#2}
	\providecommand{\BIBentrySTDinterwordspacing}{\spaceskip=0pt\relax}
	\providecommand{\BIBentryALTinterwordstretchfactor}{4}
	\providecommand{\BIBentryALTinterwordspacing}{\spaceskip=\fontdimen2\font plus
		\BIBentryALTinterwordstretchfactor\fontdimen3\font minus
		\fontdimen4\font\relax}
	\providecommand{\BIBforeignlanguage}[2]{{%
			\expandafter\ifx\csname l@#1\endcsname\relax
			\typeout{** WARNING: IEEEtran.bst: No hyphenation pattern has been}%
			\typeout{** loaded for the language `#1'. Using the pattern for}%
			\typeout{** the default language instead.}%
			\else
			\language=\csname l@#1\endcsname
			\fi
			#2}}
	\providecommand{\BIBdecl}{\relax}
	\BIBdecl
	
	\bibitem{precup2021nature}
	R.-E. Precup, E.-I. Voisan, R.-C. David, E.-L. Hedrea, E.~M. Petriu, R.-C.
	Roman, and A.-I. Szedlak-Stinean, ``Nature-inspired optimization algorithms
	for path planning and fuzzy tracking control of mobile robots,'' in
	\emph{Applied Optimization and Swarm Intelligence}.\hskip 1em plus 0.5em
	minus 0.4em\relax Springer, 2021, pp. 129--148.
	
	\bibitem{9781399}
	E.~Osaba, E.~Villar-Rodriguez, and I.~Oregi, ``A systematic literature review
	of quantum computing for routing problems,'' \emph{IEEE Access}, vol.~10, pp.
	55\,805--55\,817, 2022.
	
	\bibitem{precup2020grey}
	R.-E. Precup, E.-I. Voisan, E.~M. Petriu, M.~L. Tomescu, R.-C. David, A.-I.
	Szedlak-Stinean, and R.-C. Roman, ``Grey wolf optimizer-based approaches to
	path planning and fuzzy logic-based tracking control for mobile robots,''
	\emph{International Journal of Computers Communications \& Control}, vol.~15,
	no.~3, 2020.
	
	\bibitem{osaba2021hybrid}
	E.~Osaba, E.~Villar-Rodriguez, I.~Oregi, and A.~Moreno-Fernandez-de Leceta,
	``Hybrid quantum computing-tabu search algorithm for partitioning problems:
	Preliminary study on the traveling salesman problem,'' in \emph{2021 IEEE
		Congress on Evolutionary Computation (CEC)}.\hskip 1em plus 0.5em minus
	0.4em\relax IEEE, 2021, pp. 351--358.
	
	\bibitem{morgan2019opentripplanner}
	M.~Morgan, M.~Young, R.~Lovelace, and L.~Hama, ``Opentripplanner for r,''
	\emph{Journal of Open Source Software}, vol.~4, no.~44, p. 1926, 2019.
	
	\bibitem{huber2016calculate}
	S.~Huber and C.~Rust, ``Calculate travel time and distance with openstreetmap
	data using the open source routing machine (osrm),'' \emph{The Stata
		Journal}, vol.~16, no.~2, pp. 416--423, 2016.
	
	\bibitem{abdulrazak2022toward}
	B.~Abdulrazak, S.~Tahir, S.~Maraoui, V.~Provencher, and D.~Baillargeon,
	``Toward a trip planner adapted to older adults context: Mobila{\^\i}n{\'e}s
	project,'' in \emph{International Conference on Smart Homes and Health
		Telematics}, 2022, pp. 100--111.
	
	\bibitem{haklay2008openstreetmap}
	M.~Haklay and P.~Weber, ``Openstreetmap: User-generated street maps,''
	\emph{IEEE Pervasive computing}, vol.~7, no.~4, pp. 12--18, 2008.
	
\end{thebibliography}

\end{document}